\documentclass[letterpaper, 10 pt, conference]{ieeeconf}  % Comment this line out if you need a4paper

\IEEEoverridecommandlockouts                              % This command is only needed if 
\usepackage{graphics} % for pdf, bitmapped graphics files
\usepackage{epsfig} % for postscript graphics files
\usepackage{newtxtext,newtxmath}

\usepackage{amsmath} % assumes amsmath package installed
\usepackage{subcaption}
\usepackage{float}
\usepackage{booktabs} % For professional-looking tables
\usepackage{multirow} % For multi-row cells
\usepackage{siunitx}  % For aligning numbers by decimal point
\usepackage{xcolor}   % For coloring text
\usepackage{colortbl}  % Required for \cellcolor
\usepackage{graphicx}
\usepackage{hyperref}
\usepackage{algorithm}
\usepackage{algorithmic}

\usepackage{censor}
\title{\LARGE \bf
GS-CPE: Unified 6-Degree-of-Freedom Camera Pose Estimation via 3D Gaussian Splatting
}

\author{Huaiyuan Weng$^{1}$, Chul Min Yeum$^{1}$ and Su-Min Kang$^{2}$%  
\thanks{$^{1}$Huaiyuan Weng and Chul Min Yeum are with the Department of Civil and Environmental Engineering, University of Waterloo, 200 University Ave. W., Waterloo, ON N2L 3G1, Canada.
{\tt\small \{h22weng, cmyeum\}@uwaterloo.ca}}%
\thanks{$^{2}$Su-Min Kang is with School of Architecture, Soongsil University, 369 Sangdo-ro, Dongjak District, Seoul, South Korea {\tt\small kangsm@ssu.ac.kr}}%
\thanks{This work was supported in part by the Ontario Research Fund–Research Excellence program from the Province of Ontario under Project No. ORF-RE012-051. The views expressed herein are those of the authors and do not necessarily reflect those of the Province of Ontario. This work was also supported in part by the Brain Pool Program funded by the Ministry of Science and ICT through the National Research Foundation of Korea under Grant No. RS-2025-25401260.}
}

\begin{document}

\maketitle
\thispagestyle{empty}
\pagestyle{empty}

%%%%%%%%%%%%%%%%%%%%%%%%%%%%%%%%%%%%%%%%%%%%%%%%%%%%%%%%%%%%%%%%%%%%%%%%%%%%%%%%
% \begin{abstract}

% Although various visual localization approaches exist, such as scene coordinate regression and camera pose regression, these methods often struggle with optimization complexity or limited accuracy. We propose GS-CPE, a novel framework for accurate 6-Degree-of-Freedom (6-DoF) visual localization that combines structure-based corse pose estimation with rendering-based refinement using 3D Gaussian Splatting (3DGS). The system follows a coarse-to-fine visual localization pipeline that first performs coarse pose estimation using retrieval-based matching powered by the VGGT foundation model. It then refines the pose using differentiable rendering over 3DGS scene representation, minimizing floater-masked RGB and depth warping losses. This rendering-based optimization enables accurate and efficient 6-DoF pose refinement without the need for explicit 2D-3D correspondences. Extensive experiments across indoor and outdoor benchmarks—including 7Scenes, 12Scenes, Cambridge benchmarks, FAST-LIVO2 and custom private AC1 datasets—demonstrate state-of-the-art performance. GS-CPE outperforms existing NeRF, APR, and SCR-based localization approaches in terms of both accuracy and generalization. The project code will be open sourced.
% % The project page is available at: 
% % \href{https://anonymous.4open.science/r/GS-CPE-178C}{Anonymous GitHub Repository}. 

% \end{abstract}

\begin{abstract}
Despite substantial progress in visual localization, from scene coordinate regression to direct camera pose regression, achieving both robust generalization and high accuracy remain challenging. This study introduces GS-CPE (\underline{G}aussian \underline{S}platting based \underline{C}amera \underline{P}ose \underline{E}stimation), a coarse-to-fine framework for 6-DoF camera pose estimation that unifies geometry-based coarse pose estimation with robust 3D Gaussian Splatting (3DGS) warping-based pose refinement. GS-CPE first estimates a coarse pose via retrieval-guided geometric pose estimation on a 3DGS scene representation, then refines it by minimizing an visibility-aware masked RGB warping objective in a multi-scale optimization framework, with adaptive re-rendering. Extensive experiments on indoor and outdoor benchmarks—including 7Scenes, Cambridge Landmarks, FAST-LIVO2 datasets, and a custom dataset demonstrate state-of-the-art performance, consistently outperforming in both accuracy and generalization. Code is released at \href{https://github.com/cviss-lab/GS-CPE}{GS-CPE}.
\end{abstract}
% \begin{keywords}
% Camera Pose Estimation, Visual Localization, Gaussian Splatting
% \end{keywords}
%%%%%%%%%%%%%%%%%%%%%%%%%%%%%%%%%%%%%%%%%%%%%%%%%%%%%%%%%%%%%%%%%%%%%%%%%%%%%%%%
\section{INTRODUCTION}

Precise 6-Degree-of-Freedom (6-DoF) visual localization within a pre-built 3D map is a fundamental capability in robotics, enabling autonomous navigation, inspection, and interaction in both indoor and outdoor environments. Its applications span navigation in GPS-denied scenarios \cite{yabuuchi2021visual, sun2026view}, augmented reality \cite{ das2018joint}, and robotic inspection in structured environments \cite{lattanzi2017review}. Despite steady progress, achieving accurate localization remains challenging in practice, as real-world systems must jointly balance accuracy with robustness and efficiency, while also depending on the choice of 3D map representation.

% Owing to the low cost and widespread availability of cameras, visual localization has attracted increasing attention compared with other localization modalities. 

Classical structure-based localization methods estimate camera pose by establishing explicit 2D–3D correspondences between query keypoints and sparse map points, followed by Perspective-n-Point (PnP) with RANSAC \cite{lepetit2009epnp, fischler1981random}. Both handcrafted \cite{sattler2016efficient} and learned feature methods \cite{detone2018superpoint, sarlin2020superglue, taira2018inloc} have been proposed to improve matching robustness. Despite their broad adoption, these methods often suffer from limited accuracy because they rely on explicit feature extraction and the geometric fidelity of the 3D reconstruction.

\begin{figure}[]
    \centering
    \includegraphics[width=\linewidth]{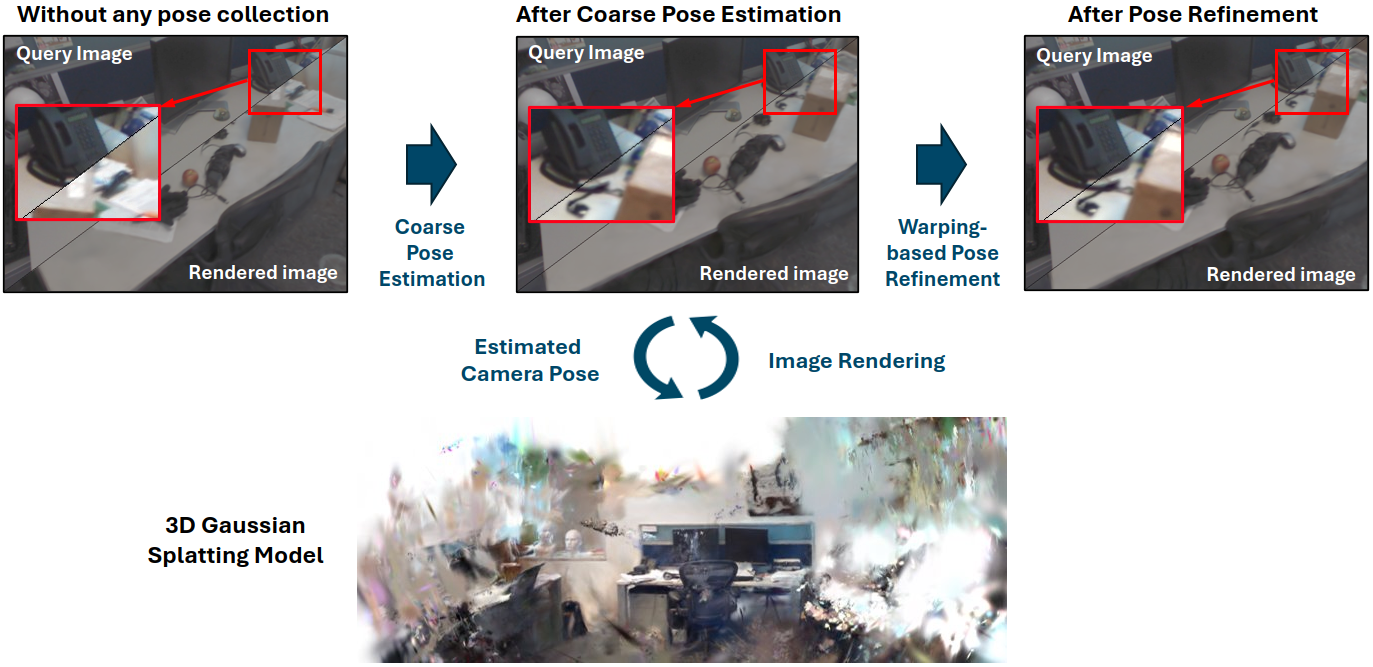}
    \caption{Teaser of \textbf{GS-CPE}. Given a query image, we estimate a coarse pose and refine it by multi-scale, visibility-masked photometric warping with adaptive re-rendering. Insets show improved alignment after refinement.}
    \label{fig:teaser}
    \vspace{-8pt}
\end{figure}

Learning-based approaches aim to overcome these limitations. Absolute Pose Regression (APR) methods directly regress camera poses from images, offering computational efficiency but often sacrificing accuracy and generalization \cite{sarlin2021back}. Scene Coordinate Regression (SCR) improves geometric consistency by predicting dense scene representations, but typically requires substantial training and optimization effort to achieve state-of-the-art performance \cite{miao2023MRL}, which may limit real-time applicability in robotics. Moreover, many learning-based methods rely on scene-specific training or adaptation, reducing practicality when scaling to new environments. 

In parallel, neural 3D scene representations such as Neural Radiance Fields (NeRF) \cite{mildenhall2021nerf} and 3DGS \cite{kerbl3Dgaussians, weng2026structuredligsstructured3dgaussians} have significantly advanced high-fidelity view synthesis, enabling Neural Render Pose (NRP) estimation. However, NeRF-based NRP commonly entails substantial computational cost during training and inference, limiting its use in real-world robotics. In contrast, 3DGS significantly improves rendering efficiency and has recently inspired a growing number of localization methods that estimate or refine poses by backpropagating through rendered views \cite{Matsuki:Murai:etal:CVPR2024, niu2024hgsloc}. However, pose refinement via differentiable rendering in 3DGS is still highly non-convex and strongly initialization-sensitive, which makes high accuracy difficult in practice.

To alleviate this issue, this study adopts warping-based optimization for pose refinement, which formulates pose refinement as an image alignment problem. Specifically, the camera pose is refined by minimizing a photometric- or feature-based error between the observed target image and a reference view warped under the current pose estimate. However, warping-based refinement remains challenging due to: (1) sensitivity to initial large viewpoint gaps make linearized warping inaccurate, leading to unreliable color/intensity transfer; (2) occlusions and visibility changes that introduce invalid pixel correspondences; and (3) optimization instability under large viewpoint or appearance gaps, where fixed linearization can easily drift.

To overcome these issues, we propose GS-CPE, a coarse-to-fine visual localization framework that unifies retrieval-initialized, geometry-based coarse pose estimation with robust 3DGS warping-based pose refinement. Given a query image, we first obtain a coarse pose through image retrieval and iterative geometry-based camera pose estimation. We then refine the pose by iteratively minimizing a visibility-aware masked RGB warping objective with 3DGS rendering updates in a multi-scale optimization framework (Fig.~\ref{fig:flowchart}). This unified design combines the global robustness of geometry-based initialization with efficient warping refinement for accurate 6-DoF localization.

Our main contributions are:
\begin{itemize}
    \item We propose a coarse-to-fine visual localization framework built on a pre-trained 3DGS scene representation, which combines retrieval-based initialization and geometry-based pose estimation.
    \item We introduce a robust warping-based pose refinement that optimizes pose by aligning 3DGS-rendered images with the query RGB image, and improves robustness via (i) adaptive re-linearization through re-rendering, (ii) multi-scale optimization, and (iii) a visibility-aware warping mask.
    \item We validate GS-CPE on indoor and outdoor benchmarks as well as additional real-world datasets, showing consistent improvements over representative APR, SCR, and NRP baselines.
\end{itemize}

\section{Related Works}

% \subsection{Geometric Pose Estimation}
% Keypoint detection and descriptor learning have evolved through deep learning, enhancing feature matching for localization. CNN-based methods, including SuperPoint \cite{detone2018superpoint}, D2-Net \cite{dusmanu2019d2}, and R2D2 \cite{revaud2019r2d2}, improve keypoint extraction and description. Transformer-based models like LoFTR \cite{sun2021loftr} SuperGlue \cite{sarlin2020superglue} and MASt3R \cite{mast3r_eccv24} further refine feature matching by capturing long-range dependencies. Feature matching-based localization methods \cite{chen2024neural} employ deep descriptors to enhance pose refinement. 

\subsection{Geometry-based Pose Estimation}
Deep learning has improved keypoint detection and description, boosting feature matching for visual localization. CNN-based methods such as SuperPoint~\cite{detone2018superpoint}, D2-Net~\cite{dusmanu2019d2}, and R2D2~\cite{revaud2019r2d2} learn robust local features, while transformer-based models like LoFTR~\cite{sun2021loftr}, SuperGlue~\cite{sarlin2020superglue}, and MASt3R~\cite{mast3r_eccv24} capture long-range context for stronger correspondences. These learned matches have been widely used for pose estimation and refinement in localization pipelines~\cite{chen2024neural}.
% In our framework, we use NetVLAD~\cite{arandjelovic2016netvlad} for retrieval-based initialization
% and employ VGGT~\cite{wang2025vggt} to obtain robust correspondences for coarse global pose estimation.

\subsection{Neural Render Pose (NRP)}
\paragraph{NeRF based NRP}
Neural Radiance Fields (NeRF) model scenes as volumetric radiance fields, encoding geometry and appearance for novel view synthesis. Methods, like iNeRF \cite{lin2021inerf}, optimize 6-DoF poses by minimizing photometric error with a frozen NeRF, while others use NeRF-rendered views for data augmentation to improve APR generalization. Direct-PoseNet \cite{chen2021directposenet} and DFNet \cite{zou2018dfnet} refine poses via photometric or feature map errors, significantly boosting localization accuracy. PNeRFLoc \cite{zhao2024pnerfloc} performs visual localization by iteratively refining camera poses using warping-based photometric consistency, but its reliance on computationally expensive NeRF rendering makes it inefficient.

\paragraph{3DGS based NRP}
3DGS has recently emerged as a powerful explicit scene representation using ellipsoidal 3D Gaussians. This formulation offers both a clear spatial structure and enables fast, high-quality novel view synthesis. In the context of visual localization, 3DGS has been primarily utilized in two ways: data augmentation and pose optimization. For data augmentation, methods like 3DGS-ReLoc \cite{jiang2024re3dgs} fuses LiDAR and camera inputs to construct a joint 3DGS map, employing normalized cross-correlation and PnP-based refinement for relocalization. Other approaches~\cite{ress2025gsplatloc, liu2025gscpr} render synthetic views from 3DGS scenes to expand training or retrieval sets. 

More commonly, 3DGS has been integrated into pose estimation and refinement pipelines. GSLoc \cite{botashev2024gsloc} introduces a differentiable renderer that backpropagates pose gradients through the rendering process, enabling gradient-based optimization. HGSLoc \cite{niu2024hgsloc} proposes a lightweight, plug-and-play refinement module that enhances coarse localization results. GS-CPR~\cite{liu2025gscpr} adopts a similar rendering-based pose refinement strategy, focusing on optimization from initially estimated poses. GSplatLoc \cite{sidorov2025gsplatloc} employs a two-stage pipeline that combines dense keypoint matching with warping-based photometric refinement. GSVisLoc~\cite{khatib2025gsvisloc} proposes a generalizable 3DGS feature matching framework for coarse-to-fine 3D–2D localization without image retrieval. 
Motivated by recent progress in geometric pose estimation and warping-based localization, we propose GS-CPE, a unified pipeline that combines retrieval-initialized geometry-based pose estimation with visibility-aware multi-scale warping-based refinement and adaptive re-linearization, improving robustness to poor initialization and occlusions.
% Building on these rendering- and warping-based localization advances, we propose GS-CPE, a unified coarse-to-fine pipeline that integrates retrieval-driven global initialization with differentiable, visibility-aware warping refinement, yielding an end-to-end pose estimation framework.

% Building on these advances, we propose a hierarchical coarse-to-fine visual localization pipeline that integrates coarse pose estimation via retrieval-based rendering methods folllowing with 3DGS warping-based refinement. Our approach incorporates learning-with-matching techniques within the refinement stage, improving both localization accuracy and generalization across scenes.

\begin{figure*}[t]
    \centering
    \includegraphics[width=0.95\linewidth]{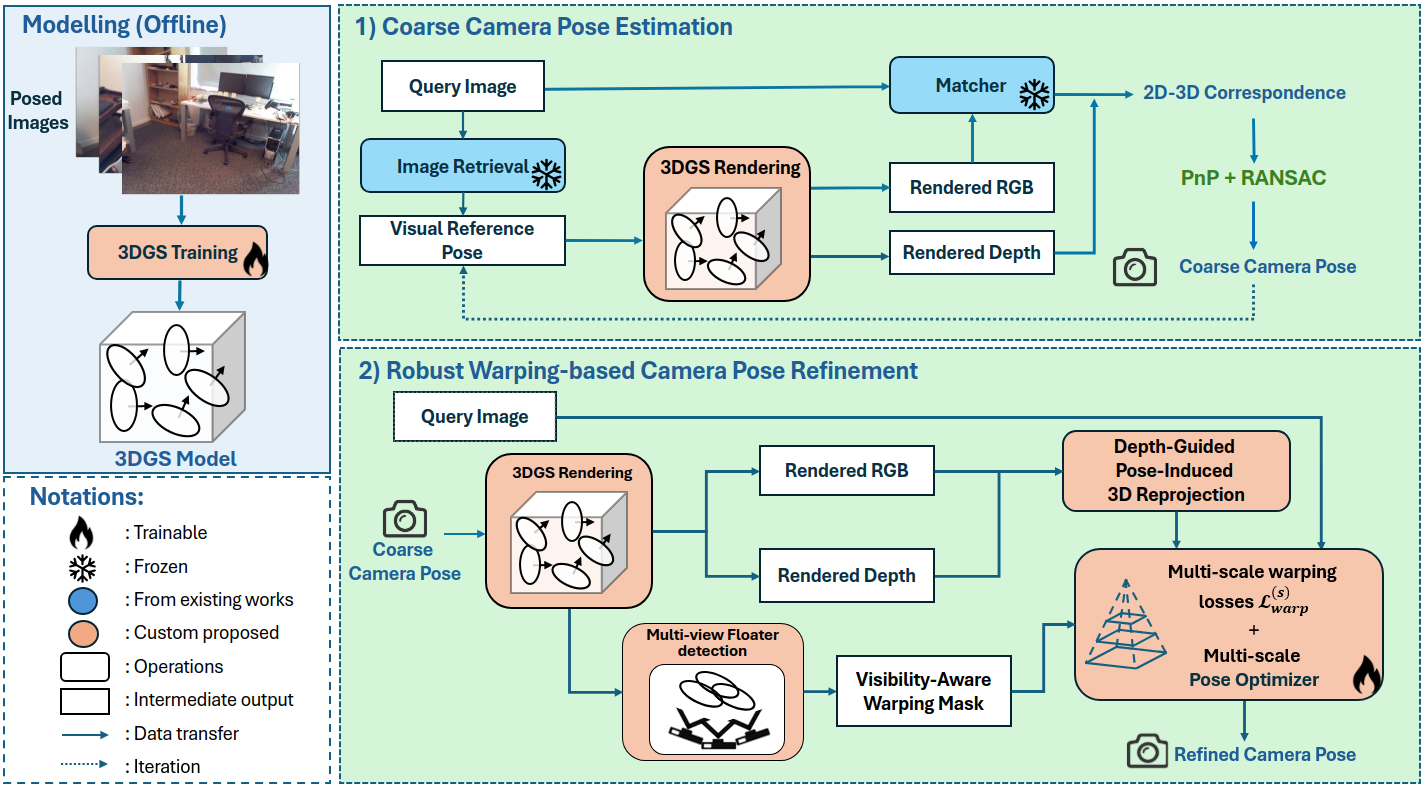}
    \caption{Overview of \textbf{GS-CPE}: The proposed method assumes that a 3DGS model is pre-built. A coarse pose is first estimated via 2D keypoint matching between the query and rendered views, followed by RANSAC-PnP. Next, pose refinement aligns the 3DGS-rendered image to the query using a visibility-aware masked RGB warping loss in a multi-scale test-time optimization.}
    \label{fig:flowchart}
\end{figure*}

% \section{Preliminaries}
% \label{sec:preliminary}
% 3DGS represents a scene as a set of $M$ anisotropic 3D Gaussians
% $\mathcal{G}=\{(\mu_i,\Sigma_i,c_i,\alpha_i)\}_{i=1}^{M}$, where $\mu_i\in\mathbb{R}^3$ denotes the mean,
% $\Sigma_i\in\mathbb{R}^{3\times 3}$ the covariance matrix, $c_i\in\mathbb{R}^3$ the RGB color, and
% $\alpha_i\in(0,1]$ the opacity \cite{kerbl3Dgaussians}. Each Gaussian defines a spatial density, and
% the collection $\mathcal{G}$ serves as an explicit and differentiable scene representation.

% Given camera intrinsics $K$ and a camera pose $T=(R,t)\in SE(3)$, the 3D Gaussians are projected onto
% the image plane and rasterized as 2D Gaussians. Through alpha blending along the viewing direction,
% this process produces a rendered RGB image and a corresponding depth map. For notational simplicity, we denote the differentiable 3DGS renderer as a function that maps the
% scene representation and camera parameters to image outputs:
% \[
% (\tilde I(T),\,\tilde D(T)) = \mathcal{R}(\mathcal{G},K,T),
% \]
% where $\tilde I(T)$ and $\tilde D(T)$ are rendered RGB and depth image, both generated from the same rendering pass under pose $T$.

\section{Preliminaries}
\label{sec:preliminary}
3DGS represents a scene as $M$ anisotropic 3D Gaussians
$\mathcal{G}=\{(\mu_i,\Sigma_i,c_i,\alpha_i)\}_{i=1}^{M}$, where $\mu_i\in\mathbb{R}^3$ denotes the mean,
$\Sigma_i\in\mathbb{R}^{3\times 3}$ denotes the covariance, $c_i\in\mathbb{R}^3$ denotes the RGB color, and
$\alpha_i\in(0,1]$ denotes the opacity~\cite{kerbl3Dgaussians}. The set $\mathcal{G}$ provides an explicit scene representation.

Given intrinsics $K$ and pose $T=(R,t)\in SE(3)$, Gaussians are projected and rasterized into 2D
Gaussians. Through alpha blending along the viewing direction, this process produces a rendered RGB and depth images. For notational simplicity, we denote the differentiable 3DGS renderer as a function that maps the
scene representation and camera parameters to image outputs:
\[
(\tilde I(T),\,\tilde D(T)) = \mathcal{R}(\mathcal{G},K,T),
\]
where $\tilde I(T)$ and $\tilde D(T)$ are rendered RGB and depth images, both generated from the same rendering pass under pose $T$.

\section{Methodology}

% We assume a pre-trained 3DGS scene representation $\mathcal{G}$ (Sec.~\ref{sec:preliminary}) and use its differentiable renderer $\mathcal{R}$. Given a query image $I_q$, our visual localization system follows a coarse-to-fine two-stage pipeline, as shown in Figure.~\ref{fig:flowchart}. In the first stage, we estimate a coarse camera pose using retrieval-based initialization followed by iterative camera pose estimation with rendered cues from $\mathcal{G}$ (Sec.~\ref{sec:corase_estimation}). In the second stage, we refine the pose by iteratively minimizing an RGB warping objective using 3DGS rendering, with custom designed strategies (Sec.~\ref{sec:fine_estimation}).

We assume a pre-trained 3DGS scene representation $\mathcal{G}$
(Sec.~\ref{sec:preliminary}) with a differentiable renderer $\mathcal{R}$. Given a query image $I_q$, our system follows a two-stage coarse-to-fine pose estimation pipeline (Fig.~\ref{fig:flowchart}). In the first stage, we estimate a coarse camera pose via retrieval-based initialization followed by iterative geometric pose estimation using cues rendered from $\mathcal{G}$ (Sec.~\ref{sec:corase_estimation}). In the second stage, we refine the pose using a robust 3DGS warping-based alignment with adaptive re-rendering and visibility-aware mask, in multi-scale optimization framework (Sec.~\ref{sec:fine_estimation}).

\subsection{Coarse Camera Pose Estimation}
\label{sec:corase_estimation}
% To estimate the coarse camera pose $T_0=(R_0,t_0)$ for a query image $I_q$, in the corase camera pose estimation phase, we estimate the camera pose through a multi-step process involving image retrieval, 2D-2D feature matching, 2D-3D correspondence establishment, and pose estimation. 

Given a query image $I_q$, we estimate an initial camera pose through a retrieval-and-geometry pipeline consisting of (1) Image Retrieval-Based Reference Initialization, (2) Rendering and 2D Feature Matching, (3) 2D--3D Correspondences, and (4) Pose estimation with PnP-RANSAC. Algorithm~\ref{alg:alg1} outlines the coarse camera pose estimation procedure.

\subsubsection{Image Retrieval-Based Reference Initialization}
\label{subsec:image_retrieval}
We first perform fast image retrieval using a retrieval operator $\mathcal{Q}$ implemented with NetVLAD~\cite{arandjelovic2016netvlad}. The query image is embedded into a global descriptor $\mathbf{f}_q$, which is compared against a database of reference images $\mathbf{I}_r=\{I_{r,j}\}_{j=1}^{N}$ with known camera poses. Based on descriptor similarity, we select the top-ranked reference image $I_{r}^{*}$ and use its associated pose $T_{r}^{*}$ as the initial pose prior for subsequent processing.

\subsubsection{Rendering and 2D Feature Matching}
\label{subsec:feature_match}
% The pose of $I_q$, denoted $T_q$, is obtained by iteratively refining $T_0$.
We initialize $T_t = T_r^{*}$, then render an RGB and depth image from
the 3DGS model $\mathcal{G}$ using the rendering function $\mathcal{R}$ defined in
Sec.~\ref{sec:preliminary}:

\begin{equation}
(\tilde I_t,\,\tilde D_t)=\mathcal{R}(\mathcal{G},K,T_t).
\end{equation}
After initialization, we iteratively update the pose $T_t$. At each iteration, we update $T_t$ and re-render $\tilde{I}_t$ to remain consistent with the current pose estimate.

We then establish 2D--2D correspondences between the query image $I_q$ and the rendered image
$\tilde I_t$ using a matcher $\mathcal{M}$. The matcher outputs a set of correspondences
\begin{equation}
\mathcal{C}_{q,t}^{(2\text{-}2)}=\{(\mathbf{X}_q,\mathbf{X}_t)\},
\end{equation}
where $\mathbf{X}_q$ and $\mathbf{X}_t$ denote matched feature locations in $I_q$ and $\tilde I_t$, respectively. In our implementation, $\mathcal{M}$ is a transformer-based matcher (VGGT\cite{wang2025vggt}) that provides robust correspondences.

\subsubsection{2D--3D Correspondences}
\label{subsec:2d3d}
% We construct query 2D--3D correspondences by combining (i) 2D--2D matches between the query image and
% the rendered reference view, and (ii) per-pixel 3D geometry obtained from the rendered depth.
% Let $C_{q,r}^{(2-2)}=\{(\mathbf{X}_q^{(2)},\mathbf{X}_r^{(2)})\}$ denote the set of 2D--2D correspondences produced by
% the matcher $\mathcal{M}$, where $\mathbf{x}_q$ is a pixel in $I_q$ and $\mathbf{x}_r$ is the matched
% pixel in the rendered image $\tilde I(T_r)$. Using the rendered depth map $\tilde D(T_r)$ and camera
% intrinsics $K$, each reference pixel $\mathbf{X}_r^{(2)}=\{(u,v)\}$ can be back-projected to a 3D point $\mathbf{X}_r^{(3)}$ in the scene:
% \begin{equation}
%     \mathbf{X}_r^{(3)} = \pi^{-1}(\mathbf{X}_r^{(2)}, \tilde D(T_r), K)
% \end{equation}
We obtain query 2D--3D correspondences by lifting the rendered-side features ($\mathbf{X}_r$) to 3D using the rendered depth ($\tilde D_t$). For each match $(\mathbf{X}_q,\mathbf{X}_t)\in\mathcal{C}_{q,t}^{(2\text{-}2)}$, we
back-project $\mathbf{X}_t=\{(u,v)\}$ with depth $\tilde D_t$:
\begin{equation}
\label{eq:backproj}
\mathbf{X}_t^{(3)}
=
\pi^{-1}\!\big(\mathbf{X}_t,\,\tilde D_t,\,K\big),
\end{equation}
where $\pi^{-1}(\cdot)$ denotes the camera back-projection operator. We then form the query 2D--3D
correspondence set
\begin{equation}
\mathcal{C}_{q, t}^{(2\text{-}3)}=\{(\mathbf{X}_q,\mathbf{X}_t^{(3)})\},
\end{equation}
which is used for pose estimation.

% where $\pi(\cdot)$ and $\pi^{-1}(\cdot)$ denote the camera projection and back-projection operators, respectively.

% We then obtain the query 2D--3D correspondence set
% $\mathcal{C}_{q}^{(2\text{-}3)}=\{(\mathbf{X}_q^{(2)},\mathbf{X}_r^{(3)})\}$ by pairing each query keypoint
% $\mathbf{X}_q^{(2)}$ with the 3D point $\mathbf{X}_r$ computed from its matched reference pixel $\mathbf{X}_r^{(2)}$. These correspondences are subsequently used for PnP with RANSAC.

\subsubsection{Pose Estimation}

Given the query 2D-3D correspondences $\mathcal{C}^{(2\text{-}3)}_{\mathrm{q, t}}$, we estimate the query
camera pose using a Perspective-$n$-Point (PnP) solver with a RANSAC loop~\cite{fischler1981random}.
Specifically, we compute
\begin{equation}
\label{eq:pnp_ransac}
T_{\mathrm{q}}
= \textsc{PnP-RANSAC}\!\left(\mathcal{C}^{(2\text{-}3)}_{\mathrm{q}},\,K,\,T_t\right)
\end{equation}
where $T_t$ provides an initial pose prior for RANSAC sampling. We then update the reference pose by setting $T_t \leftarrow T_q$, re-render
$(\tilde I_t,\tilde D_t)$, and repeat the matching and PnP steps for a fixed number of iterations or until
convergence, measured by the relative transform
$\Delta T = T_t^{-1}T_q$, i.e., $\|\Delta t\|\le \epsilon_t$ and
$\angle(\Delta R)\le \epsilon_R$, where $\Delta t$ and $\Delta R$ denote
the translational and rotational components of $\Delta T$.

% This stage is purely geometric and does not require backpropagation through either the pose solver or the 3DGS model $\mathcal{G}$. As a result, it is computationally efficient and compatible with
% standard black-box PnP implementations. Moreover, using robust 2D-3D correspondences with PnP-RANSAC provides a strong initialization for the subsequent differentiable refinement stage, while avoiding the need to train specialized local feature descriptors~\cite{chen2022dfnet, chen2024nefes}.

\begin{algorithm}
\small
\caption{Coarse Camera Pose Estimation}
\label{alg:alg1}
\begin{algorithmic}[1]
\STATE \textbf{Inputs:} database images $\mathbf{I}_{\mathrm{r}}=\{I_{\mathrm{r},j}\}_{j=1}^{N}$ with poses $\{T_{\mathrm{r},j}\}$; \\ 3DGS model $\mathcal{G}$; Image retrieval $\mathcal{Q}$; Image Matcher $\mathcal{M}$; \\ Intrinsics $K$; Convergence thresholds $(\epsilon_t,\epsilon_R)$.
\STATE \textbf{Output:} query image pose $T_{\mathrm{t}}$.
% \STATE $\mathcal{S} \leftarrow \textsc{PlaneDetect}(\mathcal{P})$ \COMMENT{Extract dominant planes from $\mathcal{P}$ (RANSAC + normal consistency)}
\WHILE{a new query image $I_{\mathrm{q}}$ arrives}
    \STATE $\mathbf{f}_{\mathrm{q}} \leftarrow \mathcal{Q}(I_{\mathrm{q}})$
    \STATE $T_{\mathrm{r}}^{*} \leftarrow \textsc{TopK}(\mathbf{f}_{\mathrm{q}},\mathbf{I}_{\mathrm{r}},1)$ \COMMENT{get pose of closest image from $\mathbf{I}_{\mathrm{r}}$} 
    \STATE $T_t \leftarrow T_{\mathrm{r}}^{*}$
        
    \FOR{iterative index $i \in \mathrm{N_c}$}
        
        \STATE $I_{\mathrm{t}}, D_{\mathrm{t}} \leftarrow \mathcal{R}(\mathcal{G},K, T_t)$ \COMMENT{rendering from $\mathcal{G}$}
        \STATE $C^{(2\text{-}2)}_{q,t}  \leftarrow \mathcal{M}(I_q, I_t)$

        \STATE $\mathbf{X}^{(3)}_{t} \leftarrow \textsc{LiftTo3D}\!\left(\mathbf{X}^{(2)}_{t},\, D_t,\, K\right)$

        % \STATE $\mathbf{X}^{(3)}_{\mathrm{q}} \leftarrow \mathbf{X}^{(3)}_{\mathrm{r}} \cup C^{(2)}_{q,r} $
        
        \STATE $C^{(2\text{-}3)}_{q,t}  \leftarrow \{\mathbf{X}^{(2)}_{\mathrm{q}}, \mathbf{X}^{(3)}_{\mathrm{t}}\}$
        
        \STATE $T_{\mathrm{q}} \leftarrow \textsc{PnP-RANSAC}(C^{(2\text{-}3)}_{q,t} ,K, T_t)$

        \STATE $\Delta T \leftarrow T_{\mathrm{t}}^{-1}T_{\mathrm{q}}$ \COMMENT{relative pose update}
        % \STATE $\Delta t \leftarrow \mathrm{trans}(\Delta T),\;\Delta R \leftarrow \mathrm{rot}(\Delta T)$
        
        \IF{$\|\mathrm{trans}(\Delta T)\|\le \epsilon_t$ \textbf{and} $\angle(\mathrm{rot}(\Delta T))\le \epsilon_R$} 
            \STATE \textbf{break} \COMMENT{exits FOR; WHILE ends next check}
        \ENDIF

        \STATE $T_{\mathrm{t}} \leftarrow T_{\mathrm{q}}$
    \ENDFOR
\ENDWHILE
\end{algorithmic}
\end{algorithm}

\subsection{Warping-based Camera Pose Refinement}
\label{sec:fine_estimation}
% Due to the limitations of geometry-based pose estimation and the uncertainty in 3DGS rendered depth, we need further refinement to improve the accuracy of camera pose. 
Inspired by warping-based refinement approaches such as PNeRFLoc~\cite{zhao_pnerfloc_2023} and
GSplatLoc~\cite{sidorov2025gsplatloc}, we employ a warping-based refinement that aligns 3DGS-rendered images with the query image $I_q$. However, warping-based refinement is sensitive to poor initialization and can be destabilized by unreliable rendered depths. To improve robustness, we incorporate: (i) adaptive re-linearization via re-rendering to reduce linearization drift \ref{subsec:adap_rendering}, (ii) a visibility-aware warping mask to suppress occlusion artifacts \ref{subsec:Visibility_mask} and (iii) multi-scale optimization for improved convergence \ref{subsec:multi-scale-opt}. Algorithm~\ref{alg:alg2} outlines the warping-based camera pose refinement.

% we iteratively refine the pose by minimizing a photometric alignment objective between the rendered image and the query image $I_q$.

\subsubsection{Basic Warping-based Pose Refinement}
The anchor pose $T_a$ is initialized from the coarse estimated pose $T_q$. We treat the anchor as a fixed reference view during refinement; therefore, the anchor is not re-rendered within a refinement stage. Specifically, the anchor image $\tilde I_a$ and depth image $\tilde D_a$ are rendered once from the 3DGS model at $T_a$:
\[
(\tilde I_a,\,\tilde D_a)=\mathcal{R}(\mathcal{G},K,T_a).
\]
During the warping-based refinement, we estimate an incremental pose update $\Delta T$ such that the refined pose is $T_f=\Delta T\,T_a$. Using the rendered depth, we align the anchor view to the query image through pose-induced reprojection.

For each pixel in the anchor image $\tilde I_a$, we first back-project it to 3D using the depth $\tilde D_a$:
\begin{equation}
    \mathbf{X}^{(3)} = \pi^{-1}(\tilde I_a,\tilde D_a, K)
\end{equation}

where $\mathbf{X}^{(3)}=\{X_j\}_{j=1}^{N}$, and then we transform it to the query camera under $T_f$, and reproject it to the image plane:
\begin{equation}
    {\mathbf{X'}}^{(2)}=\pi\!\left(K\,T_fT_a^{-1}\,\mathbf{X}^{(3)}\right)
\end{equation}
where $\pi(\cdot)$ and $\pi^{-1}(\cdot)$ denote the camera projection and back-projection operators, respectively. And $\mathbf{X'}^{(2)}=\{x'_j\}_{j=1}^{N}$ denotes the set of reprojected warp pixel locations in the query image. By construction, the index $j$ establishes a one-to-one correspondence between each anchor pixel $x_j$ and its warped location $x'_j$, forming pose-induced pixel correspondences.

We then define the photometric warping loss as
\begin{equation}
\mathcal{L}_{\mathrm{warp}}(\Delta T)
= \sum_{x}
\left\| \tilde I_a(x) - I_q\!\left(x'\right) \right\|_2^2,
\end{equation}
where $I_q(x')$ denotes the query image sampled at the reprojected location (via bilinear interpolation).

% We then define the photometric warping loss, Matched RGB colors from matched pixel location are used to build the warping loss:
% \begin{equation}
% \mathcal{L}_{\mathrm{warp}}(\Delta T)
% = \sum_{x}
% \left\| \tilde I_a(x) - I_q(x') \right\|_2^2.
% \end{equation}

We minimize $\mathcal{L}_{\mathrm{warp}}$ over $\Delta T$ using Adam optimizer:
\begin{equation}
\Delta T^{*}=\arg\min_{\Delta T}\ \mathcal{L}_{\mathrm{warp}}(\Delta T)
\end{equation}

\subsubsection{Adaptive Re-linearization via Re-rendering}
\label{subsec:adap_rendering}
The warping formulation assumes locally consistent visibility and depth around the anchor pose. When the pose update becomes large, 3D transformation will become not accurate enough for local refinement, leading to linearization drift. To mitigate this issue, we adopt adaptive re-linearization via re-rendering.

Specifically, we assign the latest refined pose as anchor pose ($T_a \leftarrow T_f$), and re-render new anchor images $(\tilde I_a,\tilde D_a)=\mathcal{R}(\mathcal{G},K,T)$
only when the accumulated pose change $\Delta T = (\Delta R, \Delta t)$ exceeds predefined thresholds on translation or rotation,
i.e.,
$\|\Delta t\| > \tilde \epsilon_t
\quad \text{or} \quad
\angle(\Delta R) > \tilde \epsilon_R$, where $\tilde \epsilon_R$ and $\tilde \epsilon_t$ denote the rotation and translation re-render threshold. By refreshing the anchor view only when necessary, this strategy preserves depth/visibility consistency for subsequent refinement while avoiding unnecessary rendering overhead.

\begin{figure}[t]
    \centering
    \includegraphics[width=0.99\linewidth]{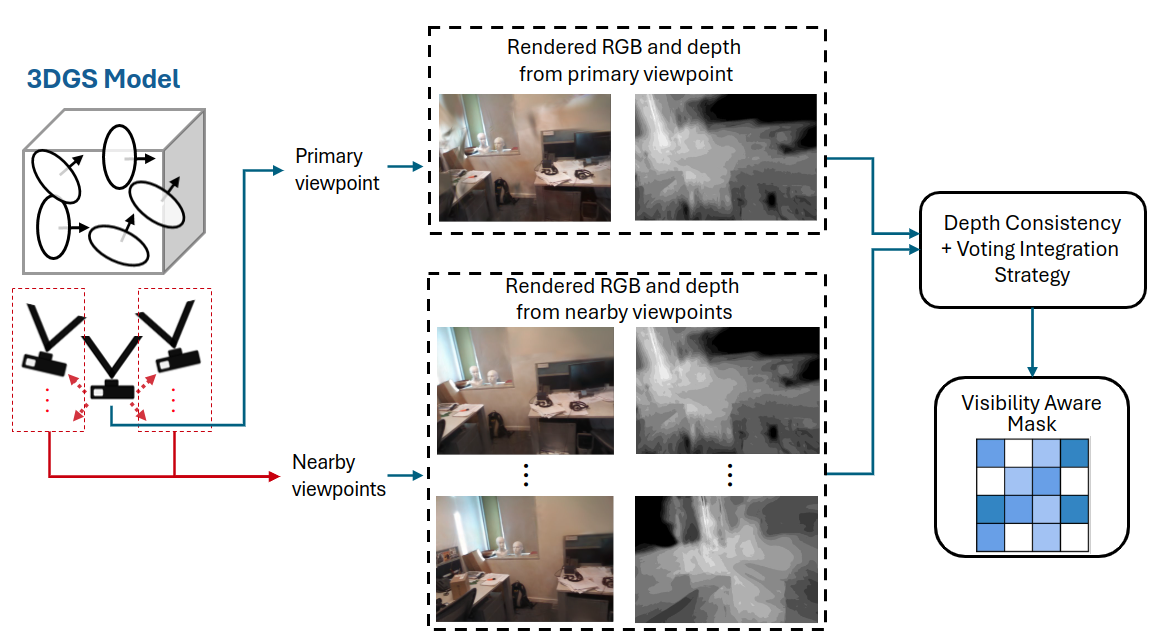}
    \caption{Visibility-aware mask generation: multi-view RGB--depth renders are converted to occupancy graphs and fused by voting to suppress depth-inconsistent regions.}
    \label{fig:floater_removal}
\vspace{-10pt}
\end{figure}

\subsubsection{Visibility-Aware Warping Mask}
\label{subsec:Visibility_mask}
We introduce a visibility-aware mask $M \in \{0, 1\}^{H \times W}$ to suppress occlusion-induced artifacts during the pose refinement. From the anchor viewpoint $T_a$, we build $V$ nearby viewpoints $\{T_v\}_{v=1}^{V}$ by applying small perturbations around the anchor pose in SE(3). We also render RGB--depth pairs $\{(\tilde I_v,\tilde D_v)\}_{v=1}^{V}$.
For anchor image $\tilde I_a$, we back-project its 3D point in the camera frame of $T_a$ as
$\mathbf{X}^{(3)}=\pi^{-1}(\tilde I_a,\tilde D_a)$. We then project it into view $v$:
\begin{equation}
Z_v^{(2)}=\pi\!\left(K\,(T_v^{-1}T_a)\,\mathbf{X}^{(3)}\right)
\end{equation}
where $Z_v^{(2)}$ is the projected depth in view $v$.
% Let $z_v^{\mathrm{proj}}=\big[(T_v^{-1}T_a)X\big]_z$ denote the projected depth in view $v$.
A view $v$ votes \emph{consistent} if $x_v$ lies inside the image domain and the rendered depth
agrees within a threshold $\tau_d$:
\begin{equation}
c_v(x)=1\!\left[ 
\left|\tilde D_v-Z_v^{(2)}\right|<\tau_d\right]
\end{equation}
We define the inconsistency score and final mask as
\begin{equation}
\vspace{-2pt}
e(x)=1-\frac{1}{V}\sum_{v=1}^{V} c_v(x)
% \vspace{-4pt}
\end{equation}
\begin{equation}
\vspace{-2pt}
M(x)=
\begin{cases}
1, & \alpha(x)>\tau_\alpha \ \text{and}\ e(x)<\tau_e,\\
0, & \text{otherwise},
\end{cases}
% M(x)=\mathbf{1}\!\left[\alpha(x)>\tau_\alpha \ \text{and}\ e(x)<\tau_e\right]
\end{equation}
where $\alpha(x)\in[0,1]$ is the per-pixel accumulated opacity of the anchor rendering, obtained by alpha-compositing the Gaussians along the ray of pixel $x$, as mentioned in Sec.~\ref{sec:preliminary}.

% We define the inconsistency score and final mask as
% \begin{equation}
% e(x)=1-\frac{1}{V}\sum_{v=1}^{V} c_v(x), \quad
% M(x)=\mathbf{1}\!\left[\alpha(x)>\tau_\alpha \ \text{and}\ e(x)<\tau_e\right].
% \end{equation}

% The masked warping loss is then formulated as:
% \begin{equation}
% \mathcal{L}_{\mathrm{warp}}(\Delta T)
% =
% \sum_{x}
% M(x)\,
% \left\|
% \tilde I_a(x) - I_q(x')
% \right\|_2^2.
% \end{equation}

\subsubsection{Multi-scale Optimization.}
\label{subsec:multi-scale-opt}
\begin{figure}[h]
    \centering
    \includegraphics[width=0.8\linewidth]{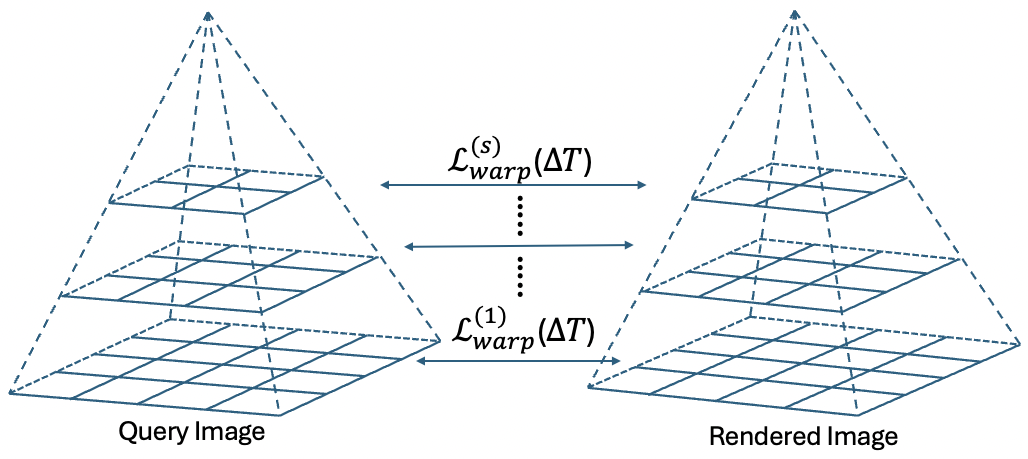}
    \caption{Multi-scale warping-based optimization: query and rendered images are processed in a multi-scale pyramid and optimized from coarse (downsampled) to the full resolution.}
    \label{fig:multi-scale}
\vspace{-8pt}
\end{figure}

As shown in Fig.~\ref{fig:multi-scale}, we construct an image pyramid with levels
$s \in \{S,\dots,0\}$, where $S$ is an integer, denoting the number of pyramid downsampling steps. In each level, both anchor images and query images are downsampled by a factor of $2^s$ to $I_a^{(s)}(x)$ and $I_q^{(s)}(x')$ respectively. At each scale, we compute a warping loss using the corresponding downsampled query and anchor images:
\begin{equation}
\mathcal{L}_{\mathrm{warp}}^{(s)}(\Delta T)
=
\sum_{x}
M(x)\,
\left\|
\tilde I_a^{(s)}(x) - I_q^{(s)}(x')
\right\|_2^2
\end{equation}

During the optimization, pose refinement proceeds from coarse to fine. Starting at $s=S$, we optimize $\Delta T$ at each level and use the resulting estimates to initialize the next finer level, repeating until $s=0$. This coarse-to-fine schedule improves convergence stability and enlarges the basin of attraction without introducing additional scale-specific weighting hyperparameters.

% Therefore, finally the warping loss is formulated as:
% \begin{equation}
% \mathcal{L}_{\mathrm{warp}}^{(s)}(\Delta T)
% =
% \sum_{x}
% M(x)\,
% \left\|
% \tilde I_a^{(s)}(x) - I_q^{(s)}(x')
% \right\|_2^2.
% \end{equation}

\begin{algorithm}[]
\caption{Warping-based Camera Pose Refinement}
\label{alg:alg2}
\small
\begin{algorithmic}[2]
\STATE \textbf{Inputs:} 3DGS model $\mathcal{G}$; Intrinsics $K$; Re-rendering thresholds $(\tilde \epsilon_t,\tilde \epsilon_R)$; Image Downsample: $\textsc{DS}$ \\
\STATE \textbf{Output:} Refined query image pose $T_{f}$.
% \STATE $\mathcal{S} \leftarrow \textsc{PlaneDetect}(\mathcal{P})$ \COMMENT{Extract dominant planes from $\mathcal{P}$ (RANSAC + normal consistency)}
\WHILE{a new query image $I_{\mathrm{q}}$ and coarse pose  $T_{\mathrm{q}}$ arrives}
    \STATE $T_a \leftarrow T_q$
    \STATE $(\tilde I_a,\,\tilde D_a)=\mathcal{R}(\mathcal{G},K,T_a)$
    \STATE $M \leftarrow \textsc{Visibility-Mask}\left( T_a, \mathcal{G} \right)$
    \FOR{scale layer $s \in \{S,\dots,0\}$}
        \STATE $I_a^{(s)}, I_q^{(s)},M^{(s)} \leftarrow \textsc{DownSample}(\tilde I_a,\,\tilde D_a,M,2^s)$
    \FOR{iterative index $i \in \mathrm{N_r}$}
        \STATE $\mathbf{X}^{(3)} = \pi^{-1}(\tilde I_a^{(s)},\tilde D_a^{(s)}, K)$ \COMMENT{2D-to-3D projection}
        \STATE ${\mathbf{X'}}^{(2)}=\pi\!\left(K\,T_fT_a^{-1}\,\mathbf{X}^{(3)}\right)$ \COMMENT{$\mathbf{X'}^{(2)}=\{x'_j\}_{j=1}^{N}$ }
        \STATE $\mathcal{L}_{\mathrm{warp}}^{(s)}(\Delta T)=\sum_{x}M^{(s)}(x) \,\left\|\tilde I_a^{(s)}(x) - I_q^{(s)}(x')\right\|_2^2$
        
        % \STATE $\Delta T \leftarrow \textsc{Adam}(\mathcal{L}_{\mathrm{warp}}(\Delta T))$

        \STATE $\Delta T^{*}=\arg\min_{\Delta T}\ \mathcal{L}_{\mathrm{warp}}^{(s)}(\Delta T)$

        \IF{$\|\mathrm{trans}(\Delta T^{*})\|\ge \tilde \epsilon_t$ \textbf{and} $\angle(\mathrm{rot}(\Delta T^{*}))\ge \tilde \epsilon_R$} 
            \STATE $T_a \leftarrow T_f \leftarrow \Delta T^{*}\,T_a$
            \STATE $(\tilde I_a,\,\tilde D_a)=\mathcal{R}(\mathcal{G},K,T_a)$ \COMMENT{re-rendering}
            \STATE $M \leftarrow \textsc{Visibility-Mask}\left( T_a, \mathcal{G} \right)$
            \STATE $I_a^{(s)}, I_q^{(s)},M^{(s)} \leftarrow \textsc{DownSample}(\tilde I_a,\,\tilde D_a,M,2^s)$
        \ENDIF
    \ENDFOR
    \ENDFOR
    \STATE $T_f \leftarrow \Delta T^{*}\,T_a$
\ENDWHILE
\end{algorithmic}
\end{algorithm}

\begin{table*}[t]
\centering
% \small
\footnotesize
\caption{Comparison of median translational and rotational errors (cm/$^\circ$) computed from $\textbf{GS-CPE}$ and existing method using 7Scenes datasets. As mentioned before, APR represents pose regression, SCR represents scene coordinate, and NRP represents for nerual render pose estimation. }
\begin{tabular}{clccccccc|c}
\toprule
\textbf{} & \textbf{Methods} & \textbf{Chess} & \textbf{Fire} & \textbf{Heads} & \textbf{Office} & \textbf{Pumpkin} & \textbf{Redkitchen} & \textbf{Stairs} & \textbf{Avg.} $\downarrow$ [cm/$^\circ$] \\
\midrule
\multirow{4}{*}{\textbf{APR}} 
  & PoseNet \cite{kendall2015posenet}  & 10/4.02 & 27/10.0 & 18/13.0 & 17/5.97 & 19/4.67 & 22/5.91 & 35/10.5 & 21.0 / 7.74 \\
  & MS-Transformer \cite{shavit2021learning}  & 11/6.38 & 23/11.5 & 13/13.0 & 18/8.14 & 17/8.42 & 16/8.92 & 29/10.3 & 18.0 / 9.51 \\
  & DFNet \cite{chen2022dfnet} & 3/1.12 & 6/2.30 & 4/2.29 & 6/1.54 & 7/1.92 & 7/1.74 & 12/2.63 & 6.0 / 1.93 \\
  & Marepo \cite{chen2024marepo} & 1.90/0.83 & 2.30/0.92 & 2.1/1.24 & 2.9/0.93 & 2.5/0.88 & 2.9/0.98 & 5.9/1.48 & 2.9 / 1.04 \\
\midrule
\multirow{3}{*}{\textbf{SCR}} 
  & DSAC* \cite{brachmann2021visual}  & 0.5/0.17 & 0.8/0.28 & \textbf{0.5}/0.34 & 1.2/0.34 & \underline{1.0}/0.28 & \textbf{0.7}/\underline{0.21} & 2.7/0.78 & 1.1 / 0.34 \\
  & ACE \cite{brachmann2023accelerated} & 0.6/0.18 & 0.9/0.34 & \underline{0.6}/\underline{0.33} & \textbf{1.0}/0.32 & 1.1/0.26 & 0.9/0.31 & 2.9/0.81 & 1.2 / 0.37 \\
  & GLACE \cite{wang2024glace} & 0.6/0.18 & 0.9/0.34 & \underline{0.6}/0.34 & 1.1/0.29 & \textbf{0.9}/\underline{0.23} & 0.8/\textbf{0.2} & 3.2/0.93 & 1.2 / 0.36 \\
\midrule
\multirow{7}{*}{\textbf{NRP}} 
  & FQN-MN \cite{germain2022feature} & 4.1/1.14 & 10.5/2.97 & 9.2/2.45 & 3.6/2.36 & 4.6/1.76 & 16.1/4.42 & 139.5/34.67 & 8.7 / 2.77 \\
  & CrossFire \cite{moreau2023crossfire} & 1/0.4 & 5/1.9 & 3/2.3 & 2/1.6 & 3/0.8 & 2/0.8 & 12/1.9 & 4.4 / 1.38 \\
  & PNeRFLoc \cite{zhao2024pnerfloc}  & 0.8/0.4 & 2.5/0.7 & 2.0/1.3 & 1.5/0.9 & 2.0/0.9 & 2.5/1.2 & 32/5.73 & 7.3/1.76 \\
  % & DFNet + NeFeS50  & 2.0/0.57 & 2.0/0.74 & 2.1/1.24 & 1.9/0.73 & 2.1/0.64 & 2.3/0.81 & 2.7/1.32 & 2.1 / 0.87 \\
  % & HR-APR  & 1.0/0.4 & 1.0/0.4 & 1.0/0.4 & 1.0/0.4 & 1.0/0.4 & 1.0/0.4 & 1.0/0.4 & 1.0 / 0.4 \\
  & NeRFMatch \cite{zhou2024nerfect} & 0.9/0.3 & 1.1/0.4 & 1.5/1.0 & 3.0/0.8 & 2.2/0.6 & 1.0/0.3 & 10.1/1.7 & 2.8 / 0.7 \\
  % & MCLoc  & 1.1/0.5 & 1.1/0.5 & 1.1/0.5 & 1.1/0.5 & 1.1/0.5 & 1.1/0.5 & 1.1/0.5 & 1.1 / 0.5 \\
  % & ACE \cite{brachmann2023accelerated}+GS-CPR \cite{liu2024gs} & 0.5/0.15 & 0.6/0.25 & 0.4/0.28 & 0.9/0.26 & 1.0/0.23 & 0.7/0.17 & 1.4/0.42 & 0.8 / 0.25 \\
  % & GSplatLoc (Coarse) \cite{sidorov_gsplatloc_2025} & 3.17/0.49 & 3.34/0.7 & 1.96/0.76 & 3.8/0.62 & 5.12/0.7 & 4.54/0.64 & 10.97/2.63 & 4.7/0.94 \\
  & GSplatLoc \cite{sidorov_gsplatloc_2025} & \textbf{0.39}/\textbf{0.13} & 0.91/0.29 & 0.94 /0.50 & 1.41/0.32 & 1.41/0.26 & 1.32/0.29 & 3.44/0.82 & 1.4/0.37 \\
  & ACE+HGSLoc \cite{niu2024hgsloc} & 0.5/0.17 & \textbf{0.6}/\underline{0.25} & \textbf{0.5}/0.29 & \textbf{1.0}/\underline{0.25} & 1.1/\textbf{0.21} & \textbf{0.7}/\textbf{0.2} & 2.8/0.69 & \underline{1.0}/\textbf{0.29}\\ 
  % & \textbf{GS-CPE (Coarse) test} &  0.83/1.0072 &  & & &  &  \\
  & \textbf{GS-CPE (Coarse)} &  0.51/0.15 & 0.97/0.32  & 0.7/0.42 & 1.28/\textbf{0.21} & 1.47/0.28 & 0.74/0.32 & \underline{1.9}/\underline{0.47} & 1.08/0.31 \\
  & \textbf{GS-CPE (Refine)} &  \underline{0.43}/\underline{0.147} & \underline{0.66}/\textbf{0.24}  & 0.67/0.52 & \underline{1.02}/\textbf{0.18} & 1.13/0.25 & \underline{0.72}/0.27 & \textbf{1.75}/\textbf{0.43} & \textbf{0.91}/\textbf{0.29}\\
  
\bottomrule
\end{tabular}

    \begin{flushleft}
    \centering
    \footnotesize * Best results are highlighted as 
    \textbf{1st} and \underline{2nd}. (lower is better)
    % and 
    % \fcolorbox{white}{yellow!30}{3rd}.
    % , and \fcolorbox{white}{yellow!30}{3rd}.
    \end{flushleft}

\label{tab:gs-cpe-7scenes}
\vspace{-12pt}
\end{table*}

\begin{table}[t]
\centering
\footnotesize
\setlength{\tabcolsep}{4pt} 
\caption{Comparison of methods across Cambridge dataset using median translational and rotational error (cm/$^\circ$).}
\begin{tabular}{clcccccccc}
\toprule
% \textbf{} & \textbf{Dataset} & \multicolumn{4}{c}{\textbf{Cambridge}}   \\
\textbf{} & \textbf{Methods} & \textbf{Kings} & \textbf{Hospital} & \textbf{Shop} & \textbf{Church}  \\
\midrule
\multirow{4}{*}{APR} 
  & PoseNet \cite{kendall2015posenet}  & 93/2.73 & 224/7.88 & 147/6.62 & 237/5.94   \\
  & MS-Transformer \cite{shavit2021learning}  & 85/1.45 & 175/2.43 & 88/3.2 & 166/4.12  \\
  & LENS \cite{moreau2022lens} & 33/0.5 & 44/0.9 & 27/1.6 & 53/1.6 \\
  & DFNet \cite{chen2022dfnet} & 73/2.37 & 200/2.98 & 67/2.21 & 137/4.02   \\
  % & Marepo  & 1.90/0.58 & 2.30/0.94 & 2.1/1.24 & 2.9/0.93  \\
\midrule
\multirow{1}{*}{SCR} 
  & ACE \cite{brachmann2023accelerated} & 29/0.38  & 31/0.61 & \textbf{5}/\underline{0.3} & 19/0.6  \\
\midrule
\multirow{3}{*}{NRP} 
  & CrossFire \cite{moreau2023crossfire}  & 47/0.7 & 43/0.7 & 20/1.2 & 39/1.4    \\
  & PNeRFLoc \cite{zhao_pnerfloc_2023} & 24/\textbf{0.29}  & 28/\textbf{0.37} & 6/\textbf{0.27} & 40/0.55  \\
  % & ACE \cite{brachmann2023accelerated} + GS-CPR & 0.5/0.15 & 0.6/0.25 & 0.5/0.34 & 1.0/0.29  \\
  & \textbf{GS-CPE (Coarse)} & \underline{23}/0.31 & 24/\underline{0.52} & \textbf{5}/0.34 & \underline{18}/\underline{0.44}   \\
  & \textbf{GS-CPE (Refine)} & \textbf{21}/\underline{0.3} & \textbf{22}/0.53 & \textbf{5}/0.32 & \textbf{16}/\textbf{0.39}   \\

\bottomrule
\end{tabular}
    \begin{flushleft}
    \centering
    \footnotesize * Best results are highlighted as 
    \textbf{1st} and \underline{2nd}. (lower is better)
    % and 
    % \fcolorbox{white}{yellow!30}{3rd}.
    % , and \fcolorbox{white}{yellow!30}{3rd}.
    \end{flushleft}

\label{tab:gs-cpe-Cambridge}
\vspace{-8pt}
\end{table}

% \begin{table}[t]
% \centering
% \footnotesize
% \setlength{\tabcolsep}{4pt} 
% \begin{tabular}{clcccccccc}
% \toprule
% \textbf{} & \textbf{Dataset} & \multicolumn{4}{c}{\textbf{ETH3D}}   \\
% \textbf{} & \textbf{Methods} & \textbf{Courtyard} & \textbf{Facade} & \textbf{Delivery}  & \textbf{Bridge}  \\
% \midrule
% \multirow{4}{*}{\textbf{APR}} 
%   & PoseNet  & 18/13.0 & 17/5.97 & 19/4.67 & 22/5.91  \\
%   & MS-Transformer   & 13/13.0 & 18/8.14 & 18/4.12 & 16/8.92  \\
%   & DFNet   & 4/2.29 & 6/1.54 & 9/1.62 & 9/1.74  \\
%   & Marepo   & 2.1/1.24 & 2.9/0.93 & 2.5/0.88 & 2.9/0.98 \\
% \midrule
% \multirow{3}{*}{\textbf{SCR}} 
%   & DSAC*   & 0.5/0.34 & 1.2/0.34 & 1.0/0.28 & 0.7/0.21  \\
%   & ACE   & 0.6/0.33 & 1.0/0.32 & 1.1/0.26 & 0.9/0.31  \\
%   & GLACE   & 0.6/0.34 & 1.1/0.29 & 0.9/0.23 & 0.8/0.20  \\
% \midrule
% \multirow{3}{*}{\textbf{NRP}} 
%   & NeRFMatch   & 1.1/0.5 & 1.1/0.5 & 1.0/0.4 & 1.1/0.4   \\
%   & MCLoc   & 1.1/0.5 & 1.1/0.5 & 1.1/0.5 & 1.1/0.5   \\
%   & ACE + GS-CPR  & 0.5/0.34 & 1.0/0.29 & 1.0/0.23 & 0.7/0.19 \\
%   & \textbf{GS-CPE (Coarse)}  & 0.5/0.34 & 1.0/0.29 & 1.0/0.23 & 0.7/0.19  \\
%   & \textbf{GS-CPE (Refine)}  & 0.5/0.34 & 1.0/0.29 & 1.0/0.23 & 0.7/0.19  \\

% \bottomrule
% \end{tabular}
% \caption{Comparison of methods across Cambridge, ETH3D, FAST-LIVO2 and private dataset AC1 datasets using translational/rotational error (cm/$^\circ$). Our method ($\textbf{GS-CPE}$) is combined with different base localizers.}
% \label{tab:gs-cpe-comparison-outdoor-dataset}
% \end{table}

\section{Experiment}
\subsection{Experiment Dataset}
\subsubsection{Benchmark Dataset}
We evaluate the performance of GS-CPE across two widely used public visual localization datasets. The 7Scenes dataset \cite{glocker2013real} comprises seven indoor scenes with volumes ranging from 1–18 $\text{m}^3$. The Cambridge Landmarks dataset \cite{kendall2015posenet} represents outdoor scenarios, characterized by challenges such as moving objects and varying lighting conditions between query and training images.

\subsubsection{Additional Real-World Datasets}
We further evaluate our method on real-world outdoor and indoor datasets. Specifically, we use the public FAST-LIVO2 dataset \cite{zheng2024fast}. We report results on two sequences, CBD Building 2 and SYSU. In addition, we introduce a private AC1 dataset collected with a Robosense AC1. The collected sequences are designed to reflect challenging real-world scenarios, including narrow indoor spaces, severe occlusions and clutter.

\subsection{Experiment Setup \& 3DGS Training}

We train one 3DGS model per scene using 3DGS~\cite{kerbl3Dgaussians} for 30{,}000 iterations with depth regularization. All experiments run on a workstation with an NVIDIA RTX A5000 GPU and an AMD Ryzen Threadripper PRO 5955WX (16-core) CPU. We split each dataset into training and testing sets using different split ratios. The training set is used to train the 3DGS model and build the retrieval database, while the test set is kept fully unseen for evaluation. Dataset-specific reconstruction sources and evaluation protocols are as follows.

\textbf{7Scenes}: We use the aligned poses from \cite{Brachmann2021ICCV} as pseudo ground truth, and follow the same training/testing sequence split as \cite{Brachmann2021ICCV}. We train the 3DGS model using the improved sparse point clouds from \cite{liu2025gscpr}.\textbf{Cambridge Landmarks}: Point clouds and initial camera poses are reconstructed with COLMAP \cite{fisher2021colmap}. To mitigate dynamic objects, we apply temporal object filtering to mask moving regions before 3DGS training. \textbf{FAST-LIVO2}: We use FAST-LIVO2\cite{zheng2024fast} to obtain point clouds and camera poses for 3DGS initialization, adopt a 1:3 test/train split. SLAM poses serve as ground truth during evaluation. \textbf{AC1 Private Dataset}: We use COLMAP~\cite{fisher2021colmap} to obtain camera poses and sparse point cloud for 3DGS initialization, adopt a 1:3 test/train split. Colmap-estimated poses are used as ground truth during evaluation. 

% For baseline comparisons, ACE~\cite{brachmann2023accelerated} is trained and evaluated on the same split, while Marepo~\cite{chen2024marepo} is evaluated using its general model with scene-specific ACE headers.

\subsection{Implementation Details}
In coarse pose estimation, we set the number of RANSAC iterations to 2000, iteration limit $N_c = 3$, convergence thresholds $\epsilon_t = 0.02$ and $\epsilon_R = 0.02^\circ$. In pose refinement, we use Adam to jointly optimize translation and rotation (quaternion parameterization) with a learning rate of $3\times10^{-4}$, iteration limit of each multi-scale level $N_f = 20$, adaptive re-rendering thresholds $\tilde{\epsilon}_t = 0.5$ and $\tilde{\epsilon}_R = 0.5^\circ$, $\tau_d = 0.02m$ and $\tau_e = 0.3$ for visibility-aware warping mask, $S = 2$ for multi-scale pose optimization.

\begin{table}[t]
\centering
\footnotesize
\setlength{\tabcolsep}{2pt}
\caption{Comparison of methods across FAST-LIVO2 and private dataset AC1 datasets using  median translational and rotational error (cm/$^\circ$) across various approaches. }
\begin{tabular}{cl|cccccccc}
\toprule
\textbf{} & \textbf{Dataset} & \multicolumn{2}{c}{\textbf{Fast-livo2}} & \multicolumn{2}{c}{\textbf{AC1}} \\
\textbf{} & \textbf{Methods} &  \textbf{CBD2} & \textbf{Sculpture} & \textbf{Lab} & \textbf{Lounge}  \\
\midrule
\multirow{1}{*}{\textbf{APR}} 
  & Marepo \cite{chen2024marepo} &  1542.88/42.29 & 141.98/3.56 & 24.85/3.29 & 24/2.49  \\
  % & MS-Transformer  &   &  &  &  &  \\
  % & DFNet  & &  &  & &   \\
  % & Marepo  & 1.90/0.58  & 2.1/1.24 & 2.9/0.93 & 2.5/0.88 & 2.9/0.98 \\
\midrule
\multirow{1}{*}{\textbf{SCR}} 
  % & DSAC*  & 0.5/0.17  & 0.5/0.34 & 1.2/0.34 & 1.0/0.28 & 0.7/0.21  \\
  & ACE \cite{brachmann2023accelerated} &  \textbf{9.4}/\textbf{0.5} & 6.8/0.2 & 0.8/0.1 & 1.3/0.2  \\
  % & GLACE  & 0.6/0.18  & 0.6/0.34 & 1.1/0.29 & 0.9/0.23 & 0.8/0.20  \\
\midrule
\multirow{2}{*}{\textbf{NRP}} 
  % & PNeRFLoc \cite{zhao_pnerfloc_2023}  &   &  &  &   \\
  & \textbf{GS-CPE} (Coarse)  & 9.71/0.569  & \underline{2.68}/\underline{0.133} & \underline{0.65}/\underline{0.103} & \underline{0.93}/\underline{0.14} \\
  % & ACE + GS-CPR & 0.5/0.15 & 0.6/0.25 & 0.5/0.34 & 1.0/0.29 & 1.0/0.23 & 0.7/0.19  \\
  & \textbf{GS-CPE} (Refine) &  \underline{9.68}/\underline{0.567} & \textbf{2.12}/\textbf{0.131} & \textbf{0.61}/\textbf{0.096} & \textbf{0.88}/\textbf{0.13}\\
  % & \textbf{GS-CPE } & 0.6/0.18  & 0.5/0.34 & 1.0/0.29 & 1.0/0.23  \\
\bottomrule
\end{tabular}
 \begin{flushleft}
    \centering
    \footnotesize * Best results are highlighted as 
    \textbf{1st} and \underline{2nd}. (lower is better)
    \end{flushleft}
\label{tab:gs-cpe-comparison-real-dataset}
\vspace{-6pt}
\end{table}

% \begin{table*}[ht]
% \centering
% \footnotesize
% \caption{Comparison on the AC1 dataset (Lab and Lounge). Accuracy is the percentage of frames within pose error thresholds (10cm/5$^\circ$, 5cm/5$^\circ$, 2cm/2$^\circ$, 1cm/1$^\circ$). Median translation/rotation errors are in cm/$^\circ$. Best results in \textbf{bold}.}
% \setlength{\tabcolsep}{3pt}
% \begin{tabular}{cl|ccccc|ccccc}
% \toprule
%  &  & \multicolumn{5}{c|}{\textbf{Lab}} 
%  & \multicolumn{5}{c}{\textbf{Lounge}} \\
% \textbf{} & \textbf{Methods} 
% & 10/5 & 5/5 & 2/2 & 1/1 & Med 
% & 10/5 & 5/5 & 2/2 & 1/1 & Med \\
% \midrule
% \textbf{APR} 
% & Marepo & 1.4 & 0.0 & 0.0 & 0.0 & 24/2.49 
% & 7.69 & 0.0 & 0.0 & 0.0 & 24.85/3.29 \\

% \midrule
% \textbf{SCR} 
% & ACE & 100.0 & 93.0 & 70.4 & 39.4 & 1.3/0.2 
% & 96.2 & 96.2 & 94.2 & 69.2 & 0.8/0.1 \\

% \midrule
% \textbf{NRP} 
% & \textbf{GS-CPE} 
% & \textbf{100} & \textbf{100} & \textbf{85.9} & \textbf{57.7} & \textbf{0.88/0.13}
% & \textbf{100.0} & \textbf{99.0} & \textbf{96.1} & \textbf{76.5} & \textbf{0.61/0.096} \\
% \bottomrule
% \end{tabular}
% \label{tab:gs-cpe-comparison-ac1}
% \end{table*}

\subsection{Performance Results on Benchmarks} 
We compare GS-CPE with representative state-of-the-art methods and report median translation
and rotation errors (cm/$^\circ$) for each scene as well as the average across scenes in
Tables~\ref{tab:gs-cpe-7scenes} and~\ref{tab:gs-cpe-Cambridge}.

\paragraph{7Scenes.}
On the indoor 7Scenes benchmark, GS-CPE achieves competitive performance against SOTA neural rendering--based localization methods. The coarse stage obtains an average error of 1.08\,cm / 0.31$^\circ$, which is further reduced to 0.91\,cm / 0.29$^\circ$ after refinement (Table~\ref{tab:gs-cpe-7scenes}). Compared with the prior NRP method
GSplatLoc (1.4\,cm / 0.37$^\circ$), GS-CPE improves translation accuracy while maintaining comparable rotational performance. While GS-CPE is not the top method in every scene, it consistently achieves first- or second-tier results, demonstrating strong generalization without per-scene training or descriptor fine-tuning.

\paragraph{Cambridge Landmarks.}
% On the outdoor Cambridge Landmarks dataset, GS-CPE also delivers consistent improvements. The coarse stage achieves 23\,cm / 0.44$^\circ$ (average across scenes), which is reduced to 21\,cm / 0.39$^\circ$ after refinement (Table~\ref{tab:gs-cpe-Cambridge}). Compared with PNeRFLoc~\cite{zhao_pnerfloc_2023}, GS-CPE improves translation accuracy while achieving
% comparable rotational performance. Our method outperforms other neural rendering–based approaches such as CrossFire and achieves competitive or superior performance compared with the scene coordinate regression method ACE. These results demonstrate that GS-CPE remains robust across diverse outdoor environments without requiring scene-specific training.

On the outdoor Cambridge Landmarks dataset, GS-CPE also delivers consistent improvements. Averaged across scenes, the coarse stage achieves 23cm / 0.44$^\circ$, which is reduced to 21cm / 0.39$^\circ$ after refinement (Table~\ref{tab:gs-cpe-Cambridge}). Compared with other NPR or APR methods, GS-CPE improves translation accuracy while achieving comparable rotational performance. We note that the absolute errors on Cambridge Landmarks are generally higher than those on 7Scenes because Cambridge contains larger-scale outdoor scenes with stronger viewpoint, illumination, and dynamic changes, making both coarse retrieval and local photometric refinement more challenging.

\begin{table}[t]
\centering
\footnotesize
% \scriptsize
\caption{Comparison of methods on \textbf{Lounge} Sequence of AC1 Custom Dataset. Accuracy is reported as the percentage of frames within pose thresholds (10cm/5$^\circ$, 5cm/5$^\circ$, 2cm/2$^\circ$, 1cm/1$^\circ$), along with median translation/rotation errors (cm/$^\circ$). Best results are shown in \textbf{bold}.}
\setlength{\tabcolsep}{2pt} 
\begin{tabular}{cl|cccc|c}
\toprule
\textbf{} & \textbf{} & 10cm/5$^\circ$ & 5cm/5$^\circ$ & 2cm/2$^\circ$ & 1cm/1$^\circ$ & Median Error \\
\textbf{} & \textbf{Methods} &  $\uparrow$($\%$) & $\uparrow$($\%$) & $\uparrow$($\%$) & $\uparrow$($\%$) & $\uparrow$[cm/$^\circ$]  \\
\midrule
\multirow{1}{*}{\textbf{APR}} 
  & Marepo \cite{chen2024marepo} &    7.69 & 0.0 & 0.0 & 0.0 & 24.85/3.29   \\

\midrule
\multirow{1}{*}{\textbf{SCR}} 
  & ACE \cite{brachmann2023accelerated} &    96.2 & 96.2 & 94.2 & 69.2 & 0.8/0.1  \\
\midrule
\multirow{1}{*}{\textbf{NRP}} 
  % & \textbf{GS-CPE (Coarse)}  &   &  &  & 0.93/0.14 \\
  & \textbf{GS-CPE} & \textbf{100.0} & \textbf{99.0} & \textbf{96.1} & \textbf{76.5} & \textbf{0.61/0.096}\\
\bottomrule
\end{tabular}
\label{tab:gs-cpe-comparison-ac1-lounge}
\vspace{-8pt}
\end{table}

\subsection{Performance Results in Public Datasets} 
% Table~\ref{tab:gs-cpe-comparison-real-dataset} compares camera pose accuracy (translation/rotation errors) on FAST-LIVO2 and AC1 across representative absolute pose regression (APR), structure-based localization (SCR), and neural rendering-based pose estimation (NRP) methods. GS-CPE achieves strong and consistent performance across scenes. In particular, GS-CPE (Refine) yields the lowest errors on all AC1 scenes and is competitive on FAST-LIVO2, closely matching the best SCR method (ACE). The main exception is the CBD2 sequence, where ACE outperforms GS-CPE, likely due to a lower-quality 3DGS model caused by inaccurate SLAM poses and point-cloud initialization. Overall, refinement consistently improves the coarse estimate, validating our hierarchical optimization.

% On the private AC1 dataset, GS-CPE significantly outperforms the APR baseline (Marepo) and matches or exceeds ACE. GS-CPE achieves the lowest median errors on Lab (0.61cm/0.096°) and Lounge (0.88cm/0.13°). Tables~\ref{tab:gs-cpe-comparison-ac1-lab} and \ref{tab:gs-cpe-comparison-ac1-lounge} further report frame-level accuracy under varying thresholds: on Lab, GS-CPE reaches 100\% at 5cm/5°, 85.9\% at 2cm/2°, and 57.7\% at 1cm/1°; on Lounge, it achieves 100\% at 10cm/5° and 76.5\% at 1cm/1°, outperforming ACE and Marepo.

% Table~\ref{tab:gs-cpe-comparison-real-dataset} reports median translation and rotation
% errors (cm/$^\circ$) on FAST-LIVO2 and the private AC1 dataset: 

\paragraph{FAST-LIVO2} 
As shown in Table~\ref{tab:gs-cpe-comparison-real-dataset} , on FAST-LIVO2 dataset, GS-CPE achieves competitive performance with respect to the SCR baseline ACE.
On the Sculpture sequence, GS-CPE achieves the best accuracy (2.12,cm / 0.131$^\circ$), improving over both the coarse stage and ACE, while on CBD2 ACE is slightly better. We attribute this gap to lower 3DGS reconstruction quality in CBD2 due to inaccurate SLAM poses and point-cloud initialization, which degrades rendering fidelity. Overall, refinement consistently improves the coarse estimate, validating our approach.

\paragraph{AC1 (Private Dataset).} As shown in Table~\ref{tab:gs-cpe-comparison-real-dataset} , on AC1 dataset, GS-CPE significantly outperforms the APR baseline (Marepo) and matches or exceeds ACE. After refinement, GS-CPE achieves the best median errors on both Lab (0.61\,cm / 0.096$^\circ$) and Lounge
(0.88\,cm / 0.13$^\circ$). We further make frame-level accuracy under strict thresholds in Lounge Sequence as summarized in
Tables~\ref{tab:gs-cpe-comparison-ac1-lounge}: on Lounge,
it achieves 100\% at 10\,cm/5$^\circ$ and 76.5\% at 1\,cm/1$^\circ$, outperforming ACE and Marepo
across thresholds.

\subsection{Runtime Analysis}

% We evaluate the per-query runtime of GS-CPE on the 7Scenes dataset using the same workstation described in Section~V-B. The runtime is averaged over all test queries and decomposed into the major processing modules. The complete GS-CPE pipeline requires \textbf{1.60781~s} per query, corresponding to an average processing rate of approximately \textbf{0.62~FPS}. Specifically, the coarse localization stage requires \textbf{1.2552~s}, while the pose refinement stage requires \textbf{0.35261~s}. Within the coarse stage, image retrieval and feature matching take approximately \textbf{0.024~s} per query. A single 3DGS rendering operation requires only \textbf{0.003~s}, while visibility-mask generation takes approximately \textbf{0.020~s} per rendered frame.

We report the per-query runtime of GS-CPE on the 7Scenes dataset using the same workstation described in Section~V-B. The runtime is averaged over all test queries and decomposed into the main processing modules. The complete pipeline requires \textbf{1.608~s} per query, corresponding to approximately \textbf{0.62~FPS}. The coarse localization and pose refinement stages take \textbf{1.255~s} and \textbf{0.353~s}, respectively. Image retrieval and feature matching require approximately \textbf{0.024~s} per query. Each 3DGS rendering operation takes only \textbf{0.003~s}, while visibility-mask generation requires approximately \textbf{0.020~s} per rendered frame.

For comparison, the NeRF-based pNeRFLoc method~\cite{zhao2024pnerfloc} requires approximately \textbf{5.5~s} per query, with NeRF rendering accounting for a substantial portion of the runtime. In contrast, GS-CPE benefits from the high rendering efficiency of 3DGS. Although the proposed method performs multiple rendering operations during coarse pose estimation and iterative refinement, the associated rendering overhead remains small. As a result, GS-CPE achieves substantially faster image relocalization while still enabling repeated rendering for pose verification and refinement.

% \begin{table}[t]
% \centering
% % \footnotesize
% % \scriptsize
% \caption{Per-query runtime breakdown of GS-CPE. Runtime is reported as mean time in milliseconds.}
% \setlength{\tabcolsep}{2pt} 
% \begin{tabular}{l|ccc|c}
% \toprule
% \textbf{} & 5cm/5$^\circ$   & 2cm/2$^\circ$  & 1cm/1$^\circ$ & Median Error \\
% \textbf{Methods}  & $\uparrow$($\%$) & $\uparrow$($\%$) & $\uparrow$($\%$) & $\uparrow$[cm/$^\circ$]  \\
% \midrule
% \text{w/o multi-scale}  & 100 & 97.5 & 34.2 & 1.08/0.197  \\

% \midrule
% \text{w/o artificial mask} & 100& 98.6  & 37.8 & 1.14/0.191  \\
% \midrule
% \text{full} & 100 & \textbf{99}  & \textbf{43.0} & \textbf{1.02}/\textbf{0.184}\\
% \bottomrule
% \end{tabular}
% \label{tab:ablation}
% \end{table}

\subsection{Ablation}
We conduct an ablation study in Table.~\ref{tab:ablation} to assess the impact of
(1) multi-scale warping-based optimization and (2) the visibility-aware consistency mask.
Starting from a single-scale baseline, we incrementally enable each component. Multi-scale optimization improves convergence robustness, while the visibility-aware mask
reduces occlusion-induced errors. Their combination achieves the best overall performance,
highlighting their complementary contributions.

\begin{table}[t]
\centering
% \footnotesize
% \scriptsize
\caption{Ablation Study on Office Sequence of 7Scenes dataset. The table reports the percentage of frames within different position/orientation thresholds and the median translation/rotation errors (cm/°). Best results are in \textbf{bold}.}
\setlength{\tabcolsep}{2pt} 
\begin{tabular}{l|ccc|c}
\toprule
\textbf{} & 5cm/5$^\circ$   & 2cm/2$^\circ$  & 1cm/1$^\circ$ & Median Error \\
\textbf{Methods}  & $\uparrow$($\%$) & $\uparrow$($\%$) & $\uparrow$($\%$) & $\uparrow$[cm/$^\circ$]  \\
\midrule
\text{w/o multi-scale}  & 100 & 97.5 & 34.2 & 1.08/0.197  \\

\midrule
\text{w/o artificial mask} & 100& 98.6  & 37.8 & 1.14/0.191  \\
\midrule
\text{full} & 100 & \textbf{99}  & \textbf{43.0} & \textbf{1.02}/\textbf{0.184}\\
\bottomrule
\end{tabular}
\label{tab:ablation}
\vspace{-8pt}
\end{table}

\section{Conclusion and Future Work}
We presented GS-CPE, a unified coarse-to-fine 3DGS-based visual localization framework that combines retrieval-initialized geometry-based pose estimation from 2D--3D correspondences with robust warping-based refinement. By unifying structure-based matching with efficient 3DGS synthesis and test-time pose refinement, GS-CPE achieves improved accuracy and practical efficiency. Experiments on benchmark and datasets demonstrate consistent gains over representative APR, SCR, and NeRF-based NRP baselines. For future work, we will extend GS-CPE to Gaussian Splatting models built with different Gaussian construction algorithms, evaluate scalability in larger real-world scenes, and deploy the system in real robotic applications.


%%%%%%%%%%%%%%%%%%%%%%%%%%%%%%%%%%%%%%%%%%%%%%%%%%%%%%%%%%%%%%%%%%%%%%%%%%%%%%%%

%%%%%%%%%%%%%%%%%%%%%%%%%%%%%%%%%%%%%%%%%%%%%%%%%%%%%%%%%%%%%%%%%%%%%%%%%%%%%%%%

%%%%%%%%%%%%%%%%%%%%%%%%%%%%%%%%%%%%%%%%%%%%%%%%%%%%%%%%%%%%%%%%%%%%%%%%%%%%%%%%

% \section*{ACKNOWLEDGMENT}
% The authors would like to take this opportunity to thank RoboSense (Suteng Innovation Technology Co., Ltd) for providing the AC1 (Active Camera) sensor.

% \addtolength{\textheight}{-12cm}
% \newpage

\bibliography{citations}
\bibliographystyle{ieeetr}
\end{document}